\documentclass[10pt,twocolumn,letterpaper]{article}

\usepackage[pagenumbers]{wacv} 

\usepackage{graphicx}
\usepackage{amsmath}
\usepackage{amssymb}
\usepackage{booktabs}
\usepackage{verbatim}
\usepackage{gensymb}
\usepackage{pifont}

\usepackage[pagebackref,breaklinks,colorlinks]{hyperref}

\usepackage[capitalize]{cleveref}
\crefname{section}{Sec.}{Secs.}
\Crefname{section}{Section}{Sections}
\Crefname{table}{Table}{Tables}
\crefname{table}{Tab.}{Tabs.}

\def\wacvPaperID{1941} 
\def\confName{WACV}
\def\confYear{2024}

\begin{document}

\title{Fake It Without Making It:  \\ Conditioned Face Generation for Accurate 3D Face Reconstruction}

\author{Will Rowan$^1$, Patrik Huber$^1$, Nick Pears$^1$, Andrew Keeling$^2$ \\
\textsuperscript{1}University of York, \textsuperscript{2}University of Leeds}

\maketitle

\begin{abstract}
   Accurate 3D face reconstruction from 2D images is an enabling technology with applications in healthcare, security, and creative industries. However, current state-of-the-art methods either rely on supervised training with very limited 3D data or self-supervised training with 2D image data. To bridge this gap, we present a method to generate a large-scale synthesised dataset of 250K photorealistic images and their corresponding shape parameters and depth maps, which we call SynthFace. Our synthesis method conditions Stable Diffusion on depth maps sampled from the FLAME 3D Morphable Model (3DMM) of the human face, allowing us to generate a diverse set of shape-consistent facial images that is designed to be balanced in race and gender. We further propose ControlFace, a deep neural network, trained on SynthFace, which achieves competitive performance on the NoW benchmark, without requiring 3D supervision or manual 3D asset creation. The complete SynthFace dataset will be made publicly available upon publication.
\end{abstract}

\section{Introduction}
\label{sec:intro}

Supervised approaches for 3D face reconstruction are limited by a lack of 3D data; 3D capture is costly and time consuming, often making large-scale 3D datasets infeasible. This has led to the wide use of self-supervised approaches \cite{tewari2017mofa, tewari2018self, wu2019mvf, shang2020self, chen2020self}. However, these approaches perform poorly in metric reconstruction \cite{sanyal2019learningNoW}.



Another approach is synthesising 3D face datasets using computer graphics. Wood \etal \cite{wood2021faketill} render a large scale dataset using a parametric face model and library of hand-crafted assets to train an accurate 2D landmark regressor. They are then able to fit a 3D face model to the predicted landmarks \cite{wood20223d}. This leads to robust performance but there remains a large domain gap; the images are not photorealistic, the process requires crafted assets, and it is computationally expensive. They propose to `fake it till you make it' with crafted `fake' data enabling them to `make it' with strong performance in the real world. We `fake' it without having to make any hand-crafted assets at all.

  \begin{figure}[t!]
      \centering
      \includegraphics[width=\columnwidth]{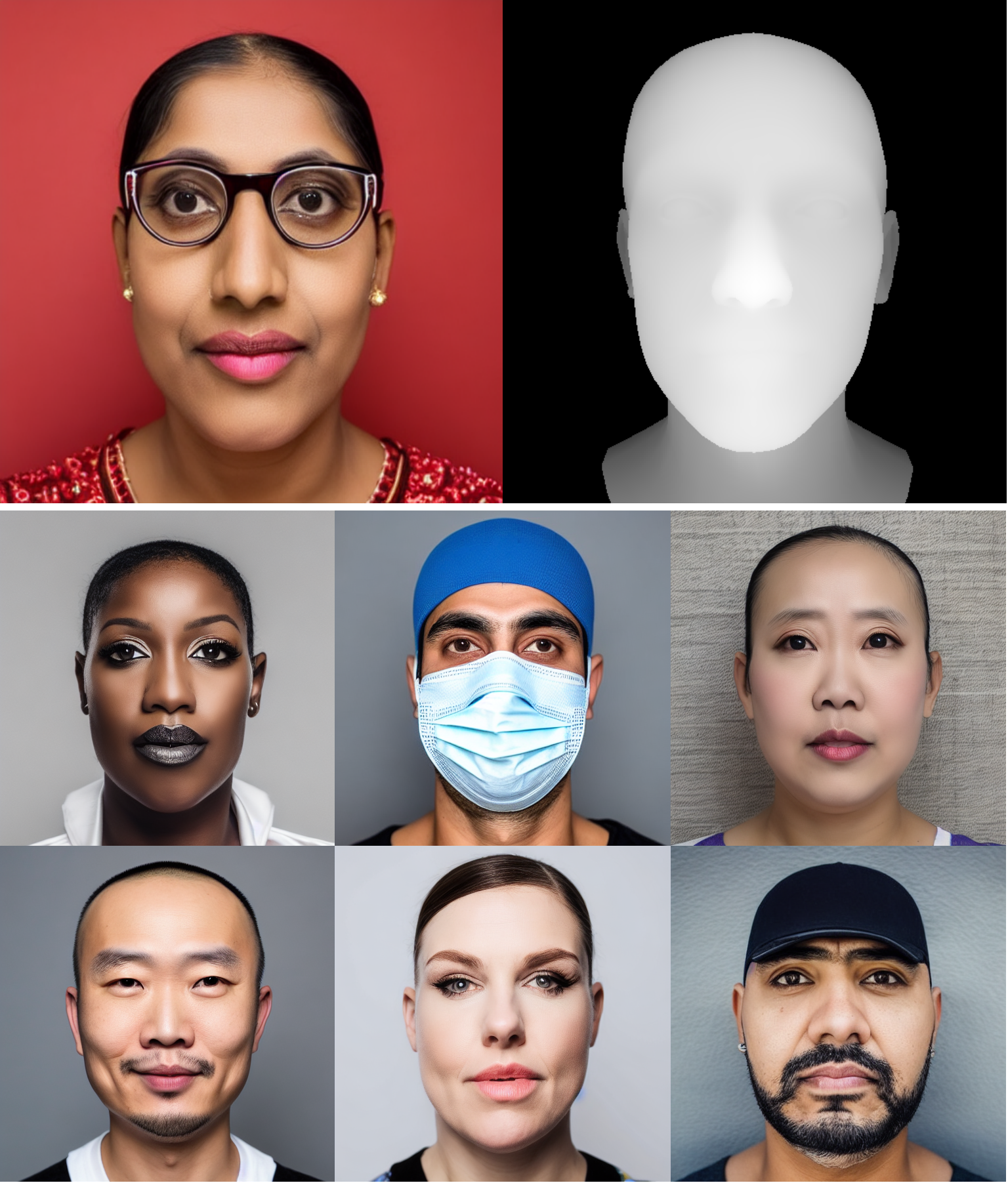}
      \caption{SynthFace, our dataset of photorealistic faces and corresponding 3D Morphable Model (3DMM) shape parameters and depth maps, is generated using conditioned Stable Diffusion and rendered depth maps from the FLAME 3DMM of the head. The first example is shown alongside the conditioning depth map. 
}
      \label{fig:SynthFace_Dataset}
  \end{figure}

      

 Zielonka \etal \cite{zielonka2022towardsMICA} annotate and unify existing 3D face datasets to enable supervised training of their MICA (MetrIC fAce) network. This is the current state-of-the-art in metric 3D face reconstruction from a single image on the NoW benchmark \cite{sanyal2019learningNoW}. However, their work represents an upper bound on a dataset for supervised 3DMM regression until more 3D data is collected.


 We overcome this limitation by devising a synthetic dataset generation pipeline that combines 2D and 3D generative models. We use the generative capabilities of a 3D Morphable Model (3DMM) as conditioning for a stable diffusion image generation model. This is achieved using ControlNet \cite{zhang2023addingcontrol}, which adds conditional control to Stable Diffusion \cite{rombach2022highstable}. We use ControlNet to condition \emph{Stable Diffusion 1.5} on depth maps of our generated 3D faces. This results in  a photorealistic 2D face image with known 3D shape, as shown in \cref{fig:SynthFace_Dataset}.

 We employed this approach to generate \emph{SynthFace}, the largest available dataset of 2D face images that has known underlying shape via their 3DMM parameters. Moreover, SynthFace is the first large-scale dataset for the supervised training of 3D face shape reconstruction networks that is balanced by both race and gender. This is achieved through linking 2D image data on the appearance of human faces with generative 3D shape information.


Commercial gender classification systems have been found to exhibit large variations in performance based on an individual’s skin tone; they misclassified dark-skinned females more than any other group and were found to be trained on datasets including predominantly light-skinned training subjects \cite{buolamwini2018gender}. Work since has shown that balancing by race and gender leads to improved performance for this task across subgroups \cite{karkkainen2021fairface}. SynthFace aims to mitigate bias using a diffusion model to balance by race and gender.




The main contributions of our work are twofold: First, we introduce SynthFace, the first large-scale synthesised dataset of 250K photorealistic faces, depth maps, and corresponding 3DMM parameters, which significantly expands the available data for training and evaluating 3D face reconstruction models. This dataset is balanced by race and gender with the aim of reducing racial and gender bias in 3D face reconstruction for trained models, in addition to enabling evaluation of other models on defined subgroups. Second, we introduce \emph{ControlFace}, a network trained on this dataset. With ControlFace, we demonstrate  competitive performance on the NoW benchmark, demonstrating that by integrating information from 2D and 3D generative models, we can improve 3D face reconstruction. In producing the first large-scale race-balanced dataset for this task, we also demonstrate how diffusion models may be utilised to improve the diversity of synthesised datasets. The complete SynthFace dataset will be made publicly available upon publication of this paper.


In summary, our work presents a novel approach to bridge the gap between the limited availability of 3D data and the abundance of 2D data for face shape estimation. Our method is simple to implement, easily extensible, and computationally inexpensive. Future improvements to image generation models, conditioning methods, and 3D face models can all be easily exploited using our method. Through introducing SynthFace and demonstrating the effectiveness of ControlFace, we reveal a promising new direction for improving 3D face shape estimation.

\section{Related Work}

Three-dimensional face reconstruction from a single image represents a significant challenge in the field of computer vision. It is an ill-posed problem due to the effects of perspective and scaling.
To tackle this, 3D Morphable Models (3DMMs) have been extensively used since their introduction  by Blanz and Vetter \cite{blanz1999morphable}, as they offer prior knowledge of human facial structure to help resolve ambiguities.
3DMMs provide a compact representation of the human face, allow additional constraints to be placed on reconstructions, and facilitate morphing between faces. Furthermore, their generative capabilities enable the sampling of realistic, geometrically consistent faces from within the model's space \cite{egger20203d}.

However, despite the widespread success of supervised learning across computer vision tasks, it has been severely limited in 3D face reconstruction due to a lack of training data. In this context, supervised learning involves the use of paired 2D-to-3D data, whether real or synthetic, which formally comprises a set of face images and their corresponding 3D model representations \cite{sanyal2019learningNoW}.
To navigate the scarcity of 3D supervision, many recent approaches have considered optimisation-based and self-supervised methods, but these have shown poor performance on metric benchmarks \cite{sanyal2019learningNoW}. Consequently, there is a need to explore supervised approaches to reconstruction and the collection of large-scale 3D training data to simplify the task.

This lack of training data is even more acute for applications that rely on events that are rarely occurring, such as facial trauma. This is the case in maxillofacial prosthesis design where clinicians want to reconstruct a missing region of the face following an accident or surgical intervention. The facial areas of such patients are not modelled in standard datasets for 3D face shape estimation. An ongoing clinical trial is comparing digitally manufactured prostheses with conventional manufacture \cite{jablonski2023improving}. 3DMMs will be used in the digital arm of this trial for facial completion.

Our work in unconstrained dataset generation allows for the modelling of rarer clinical cases; for example, in the case of orbital (eye) defects. Orbital reconstruction can then benefit from a dataset designed for the task. Current landmark-based methods struggle in the presence of asymmetrical facial defects. Our method can enable learning-based approaches in the absence of sufficiently-large real datasets.

In this work, we explore how the analytical and generative applications of 3DMMs can be combined to achieve accurate 3D face reconstruction. To achieve this, we examine current supervised methods for reconstruction, photorealistic face generation in both 2D and 3D, and how these approaches can be integrated to enable accurate 3D face reconstruction.




\subsection{Supervised Reconstruction}

One of the earliest notable approaches to supervised reconstruction using deep learning is by Tran \etal \cite{tuan2017regressing}. They create surrogate ground truth parameters using pooled multi-image 3DMM estimation. This process involves optimisation-based reconstructions for each image of an individual, with final shape and texture parameters being a weighted average of individual estimations. This is a clever observation: taking advantage of existing 2D multi-image data to improve 3D reconstruction from a single image. This dataset is then used for supervised training with a deep CNN. Despite its novelty in leveraging existing 2D multi-image data for improved 3D reconstruction, this approach is inherently limited by the initial reconstruction method used to generate the training data; at best, it can learn to be as good as this method.

Richardson \etal \cite{richardson20163d} generate face geometries directly from a 3DMM, rendering the face as an image under randomised lighting conditions. This results in a dataset of images with known 3DMM parameters; however these images are far from photorealistic. This points to a wider problem in synthesised approaches: a domain gap between synthesised and real data that makes generalisation difficult and task performance poor \cite{kar2019meta}.

In contrast, Wood \etal \cite{wood2021faketill} render highly realistic 3D face models for landmark localisation, demonstrating that synthesised data can be used to solve real world problems in the wild. Wood \etal \cite{wood20223d} build upon this work to train a dense landmark regressor for 702 facial points. A morphable model is fitted to these dense landmarks, leading to state-of-the-art results in 3D face reconstruction. 

The success of this approach affirms the potential of network-based methods in advancing 3D shape estimation. However, this approach requires the manual creation of 3D assets with associated time, financial, and computational costs. Furthermore, the rendered images fall short of photorealism which limits their uses for direct 3DMM regression. 

Other approaches have considered using the 3D data we have rather than relying on synthesised datasets. Zielonka \etal \cite{zielonka2022towardsMICA} achieve state-of-the-art performance on the NoW benchmark through unifying existing 3D face datasets. This demonstrates the importance of supervision for reconstruction performance even when supervised with minimal available data. However, this approach already represents the upper bound for supervised learning using 3D data, unless further data is collected. In combining 8 existing datasets, they reach just 2315 individuals; this remains a small dataset for supervised learning techniques. Hence, a generative approach similar to Wood \etal \cite{wood2021faketill} is required for unconstrained dataset generation.

Other significant works in this field include exploring a hybrid loss function for weakly-supervised learning \cite{deng2019accurate}, generating surrogate ground truth data via multi-image 3DMM fitting using joint optimisation \cite{liu2018disentangling}, and learning an image-to-image translation network using known depth and feature maps generated from a 3DMM \cite{sela2017unrestricted}.

In our work, we build upon these existing supervised learning methods, combining 2D generative image models and 3D face models. This approach allows us to develop a dataset larger than that proposed by Wood \etal \cite{wood2021faketill} but without the extensive effort required to create 3D assets. We `fake it' without making it. By leveraging state-of-the-art generative image models, we generate photorealistic images comparable to those used to train MICA \cite{zielonka2022towardsMICA} while being able to scale our dataset size to orders of magnitude above theirs. By taking this novel approach, we aim to significantly advance the field of 3D face reconstruction, introducing a new methodology to achieve accurate 3D face reconstruction using supervised learning.


\subsection{Optimising Identity Vectors}

The loss function used for supervised 3D reconstruction requires careful consideration. Tran \etal \cite{tuan2017regressing} introduce an asymmetric Euclidean loss for minimising errors between predicted and actual parameter vectors; this decouples over-estimation errors from under-estimation errors. A standard Euclidean loss favours estimates close to 0 due to 3DMM parameters following a multivariate Gaussian distribution centred at zero by construction. They report more realistic face reconstructions using their asymmetric Euclidean loss. 

However, these losses minimise distance in the vector space of 3DMM parameters rather than minimising reconstruction error directly. Richardson \etal \cite{richardson20163d} directly calculate the Mean Squared Error (MSE) between generated 3D mesh representations. This ensures the loss takes into account how the parameter values affect the reconstructed geometry. Zielonka \etal \cite{zielonka2022towardsMICA} also employ a mesh-based loss, but they introduce a region-dependent weight mask to weigh the facial region much more heavily than the rest of the head. We aim for accurate 3D face shape estimation, so we will optimise directly in the 3D space using a mesh loss.


\subsection{Realistic Parameterised Faces}

Automating the tedious manual work behind photorealistic face generation remains an open challenge and long-term goal of 3D face representations \cite{egger20203d}. 3DMMs provide parametric control but generate unrealistic images; Generative Adversarial Networks (GANs) generate photorealistic images but lack explicit control \cite{ghosh2020gif}. Combining the parametric control of a 3DMM with the expressive power of generative image models for faces has the potential to create large-scale datasets for supervised 3D face reconstruction.


Recent work has sought to harness the best of both worlds. StyleRig \cite{tewari2020stylerig} was the first approach to offer explicit control over a pretrained StyleGAN through a 3DMM, allowing for parametric editing of generated images. Building upon this, Ghosh \etal \cite{ghosh2020gif} condition StyleGAN2 \cite{karras2020analyzing} on rendered FLAME \cite{li2017learningFLAME} geometry and photometric details to add parametric control to GAN-based face generation, facilitating full control over the image generation process. Sun \etal \cite{sun2022cgof++} propose a NeRF-based 3D face synthesis network which enforces similarity with a mesh generated by a 3DMM. However, in all these cases, the resulting images fall short of photorealism.

In the field of image synthesis, probabilistic diffusion models now represent the state-of-the-art, surpassing the capabilities of GANs \cite{dhariwal2021diffusion}. These models, which have developed significantly since their proposal \cite{sohl2015deep}, have been further improved by concurrent advances in transformer-based architectures \cite{vaswani2017attention} and text-image embedding spaces \cite{ramesh2021zero}. Publicly available text-image embedding spaces such as CLIP \cite{radford2021learning} have further diversified and enhanced these models \cite{ramesh2022hierarchical}.

Stable Diffusion is a powerful text-to-image diffusion model, synthesising high resolution images from textual prompts using a Latent Diffusion architecture \cite{rombach2022highstable}. ControlNet \cite{zhang2023addingcontrol}, a HyperNetwork that influences the weights of a larger paired network \cite{ha2016hypernetworks}, enables a variety of input modalities to be used to condition the output of Stable Diffusion. Implementations include depth maps, user sketches, and normal map conditioning networks, among others. We use the depth version of ControlNet 1.1. It utilises MiDaS \cite{ranftl2020towards} to obtain 3,000,000 depth-image-caption pairs for training.

Unlike previous methods, ControlNet enables photorealistic image generation with strong shape control. For our use case, this enables us to create our own large-scale dataset of photorealistic images and known 3DMM parameters, with conditioning depth maps being generated from an existing model of 3D face shape.



  \begin{figure*}
      \centering
      \includegraphics[width=\textwidth]{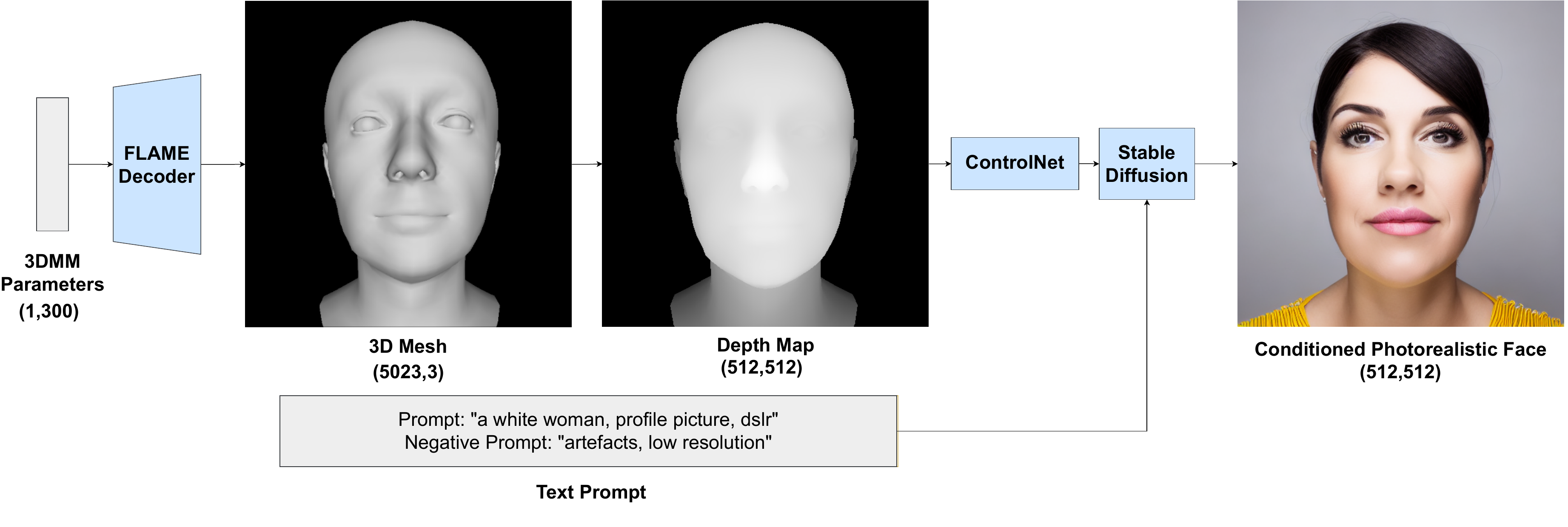}
      \caption{The SynthFace Generator. We sample from a 300-dimensional shape vector and use the FLAME decoder to produce a 3D mesh. From this mesh, we extract a depth map that, alongside a textual prompt, is used as conditioning to generate a photorealistic face.}
      \label{fig:SynthFaceGenerator}
  \end{figure*}

\section{SynthFace: Fake It Without Making It}

We present a synthetic face generator, the \emph{SynthFace Generator}, and employ it to generate a comprehensive training dataset for 3D face reconstruction, comprising 250K photorealistic faces with 10K distinct 3D facial shapes. We call this the \emph{SynthFace Dataset}. We render 250K $(512,512)$ resolution images in 40 hours utilising 10 GTX 1080 GPUs, which demonstrates an order of magnitude lower resource requirement compared to similar work \cite{wood2021faketill}. 

The generator works as follows. First, we sample 10K faces from the FLAME head model. For each of these faces, we render five depth maps under different perspective projections; this is achieved by setting a constant 72.4\degree field-of-view and varying the distance between camera and subject. This gives us 50K depth maps. There are 10K distinct 3D shapes within the SynthFace dataset but each depth map captures a different perceived shape due to the effects of perspective projection. This is designed to enable networks trained on the SynthFace dataset to disentangle identity and perspective effects from the underlying 3D shape. We then use \emph{ControlNet 1.1} to condition \emph{Stable Diffusion 1.5} to produce photorealistic faces that adhere to the shape of these depth maps. This is performed five times for each depth map, which allows networks trained with the SynthFace dataset to model the fact that different people with highly different appearances in terms of colour-texture can have the same (or highly similar) underlying face shape. This results in 250K photorealistic images with corresponding 3DMM shape parameters and depth maps. \Cref{fig:SynthFaceGenerator} shows this pipeline.


We use textual prompts to create a race-balanced dataset. Following the work of K\"{a}rkk\"{a}inen and Joo \cite{karkkainen2021fairface}, we balance our dataset to include equal representation of seven defined race groups: White, Black, Indian, East Asian, Southeast Asian, Middle Eastern, and Latino. 
(Section 1 in the supplementary material includes a full explanation of the prompts used to create SynthFace and exploration of the generated images.) To the best of our knowledge, SynthFace is the first large-scale race-balanced dataset for supervised 3D reconstruction.

We further include three common types of occlusion within the SynthFace dataset: glasses, sunglasses, and face coverings. Details such as facial hair, wrinkles, and hair are captured within the diffusion process without specific specification. This adds further realism and introduces additional occlusions into our dataset. We do not model hair present outside the defined shape of the depth maps. ArcFace is able to extract a face descriptor in the presence or absence of further hair, enabling our approach to work irrespective of this. We leave further modelling of hair to future work. 

\begin{figure}
      \centering
      \includegraphics[width=\columnwidth]{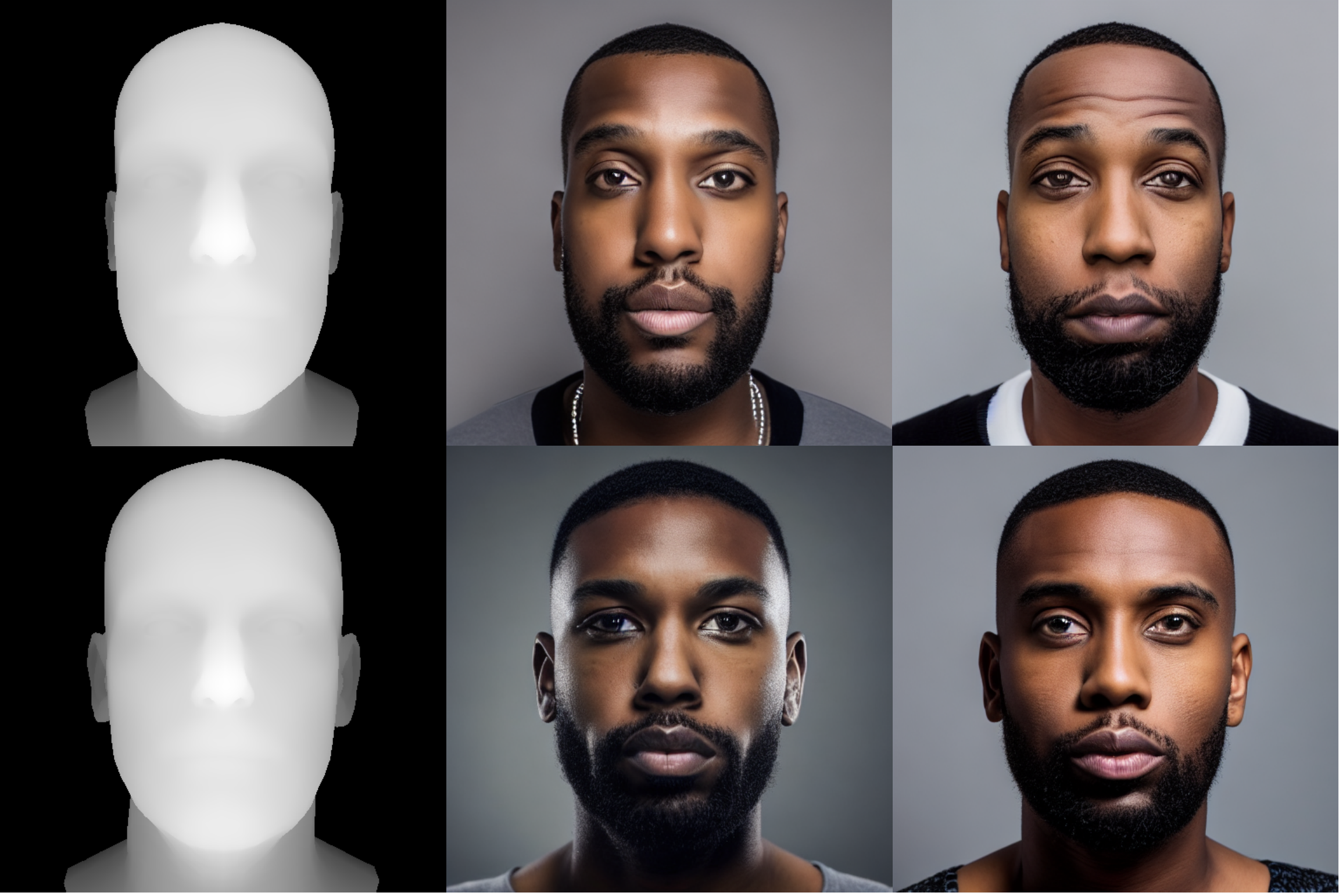}
      \caption{The SynthFace Dataset includes different perspective projections and visual identities for the same 3D shape. The first column displays two rendered depth maps of the same 3D shape but under different perspective projections. The following images in each row are conditioned on that depth map.}
      \label{fig:SynthFace_shape_exploration}
  \end{figure}

In contrast to other 3D face datasets, we include a large number of different identities for the same face shape. An identity here is an individual recognisable person in 2D image space; a shape is the 3D mesh as parameterised by the 3DMM. We produce 25 images per distinct 3D shape, each capturing a different visual identity, but with the same underlying 3D shape. \Cref{fig:SynthFace_shape_exploration} shows how different identities are included within SynthFace for the same shape. We believe we are the first to incorporate this approach into a dataset for 3D face shape estimation by design. Hence, SynthFace enables disentanglement of shape and identity through supervised learning.

\subsection{3D Face Model}

We use the FLAME head model \cite{li2017learningFLAME} as a generative model for face shape. FLAME is a linear 3DMM with both identity and expression parameters. Linear blend skinning (LBS) and pose-determined corrective blendshapes are used to model the neck, jaw, and eyeballs around joints. This results in a head model containing N = 5023 vertices and K = 4 joints. 
FLAME takes coefficients for shape $\vec{\beta} \in \mathbb{R}^{|\beta|},$ pose $ \vec{\theta} \in \mathbb{R}^{|\theta|},$ and expression $ \vec{\psi} \in \mathbb{R}^{|\psi|}$. These are modelled as vertex displacements from a template mesh $\mathbf{\overline{T}}$. A skinning function $W$ rotates the  vertices of $T$ around joints $J \in \mathbb{R}^{3K}$. This is linearly smoothed by blendweights $\mathcal{W} \in \mathbb{R}^{K \times N}$. The model is formally defined as:

\begin{equation}
M(\vec{\beta}, \vec{\theta}, \vec{\psi}) = W(T_P(\vec{\beta}, \vec{\theta}, \vec{\psi}), \mathbf{J}(\vec{\beta}), \vec{\theta}, \mathcal{W}) \tag{1}
\label{FLAME_1}
\end{equation}
where

\begin{equation}
T_P(\vec{\beta}, \vec{\theta}, \vec{\psi}) = \mathbf{\overline{T}} + B_S(\vec{\beta}; S) + B_P(\vec{\theta}; P) + B_E(\vec{\psi}; E). \tag{2}
\label{FLAME_2}
\end{equation}

Due to different face shapes requiring different joint locations, joints are defined as a function of $\vec{\beta}$. \Cref{FLAME_2} includes shape, pose, and expression blendshapes. We sample shape coefficients and set pose and expression coefficients to 0. We use \cref{FLAME_1} to generate a complete 3D mesh of the head from these coefficients.


This approach enables us to create an arbitrary number of human head shapes, each compactly represented by a set of 3DMM parameters. Approaches which directly render textured versions of meshes to 2D suffer from low-fidelity, unrealistic outputs. Instead, we extract the depth map of each mesh to pass to ControlNet, generating realistic faces in the 2D domain. 



\subsection{Depth Map Generation}


In building the SynthFace dataset, we use all 300 FLAME shape parameters ($\vec{\beta}$). We later use ArcFace as a feature extractor \cite{deng2019arcface}. ArcFace uses a novel additive angular margin loss to increase inter-class distance while reducing intra-class distance. This network has been trained to extract discriminative facial features with invariance to rotation and expression of the face for recognition tasks. Hence, we chose not to model these variations within our dataset. We believe this learning is better performed in the 2D domain with pretrained networks specialised for these tasks.


We sample identity parameters, $\vec{\beta}$, individually from a Gaussian distribution with mean 0 and s.d. 0.8. This enables a wide variation of face shape within our dataset. Expression coefficients, $\vec{\psi}$, are set to 0. We further set pose coefficients, $\vec{\theta}$, to 0. This results in a fixed frontal pose, which is suitable as input for identity descriptor networks such as ArcFace \cite{deng2019arcface}. We use a perspective camera with a 72.4\degree field of view. We vary the distance between the camera and subject from 150 to 400 world units using uniform sampling. This leads to perspective projection effects that model real-world image changes, enabling network training to learn to deal with these effects.

  \begin{figure}
      \centering
      \includegraphics[width=\columnwidth]{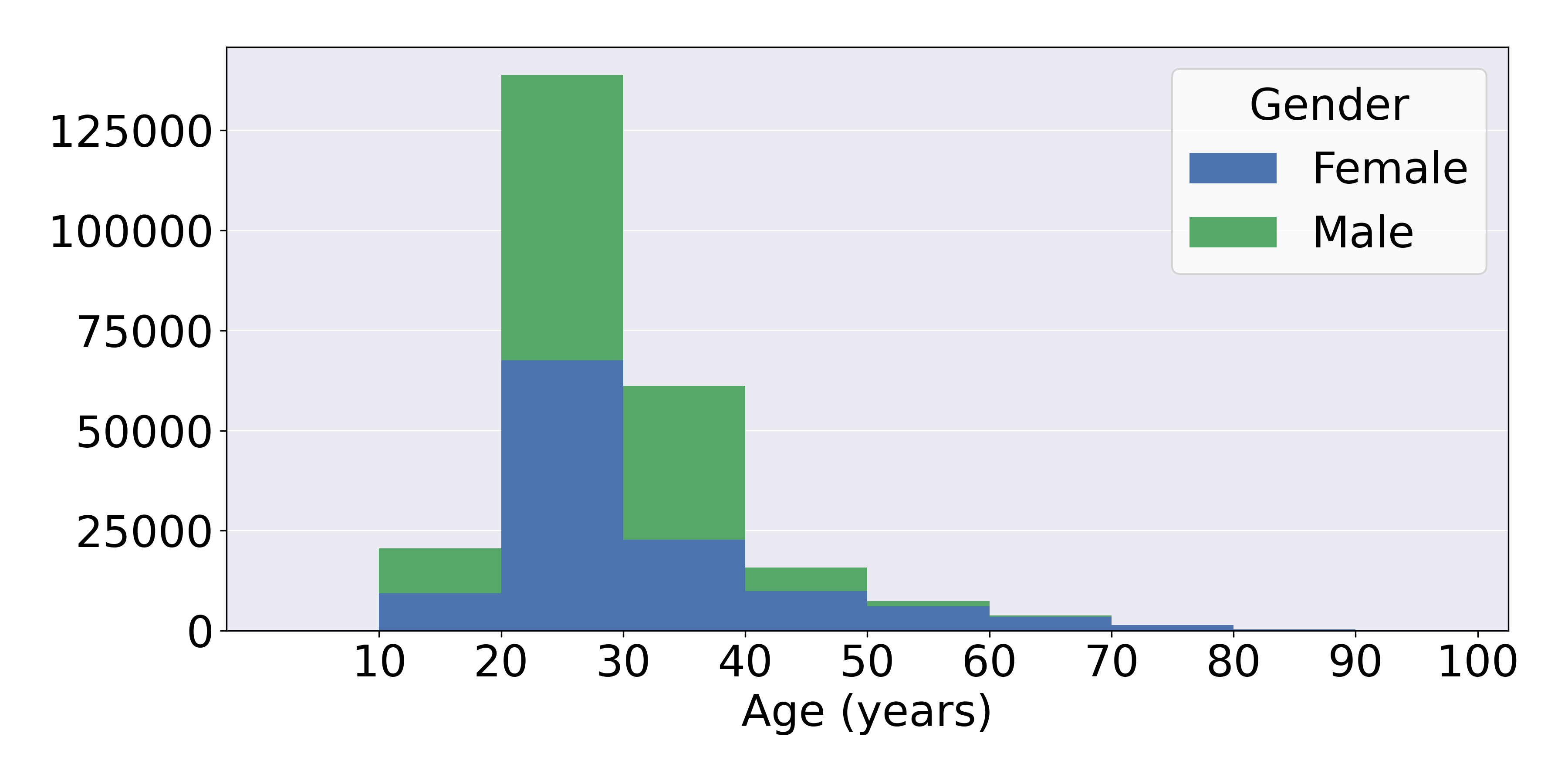}
      \caption{The SynthFace Dataset age distribution.}
      \label{fig:SynthFace_age_distribution}
  \end{figure}

\subsection{Conditioned Face Generation}

We use the depth version of \emph{ControlNet} to modulate the output of \emph{Stable Diffusion 1.5}. It takes a depth map and textual prompts (positive and negative prompts) as input to produce an image. We produce 5 images per prompt. The inference procedure is set to run for 15 steps. We use customised prompts for race, gender, and the three main types of occlusions. This results in the following prompt skeleton: ``$\left\{occlusion\right\}$, $\left\{race\right\}$ $\left\{gender\right\}$, studio portrait, profile picture, dslr". Negative Prompt: ``artefacts, low resolution". 30\% of all images in the SynthFace dataset model occlusions. This is split equally between glasses, sunglasses, and face masks. The prompts for these are given in \cref{eq:occlusions}. All images, including those under occlusion, are split equally by race and gender as defined in \cref{eq:race} and \cref{eq:gender} respectively.

\begin{align}
\label{eq:occlusions}
\text{occlusions} &= \left\{ 
    \begin{array}{l}
        \text{glasses, sunglasses,} \\
        \text{surgical mask covering face}
    \end{array} 
\right\} \\
\label{eq:race}
\text{race} &= \left\{ 
    \begin{array}{l}
        \text{White, Black, Indian,} \\
        \text{East Asian, Southeast Asian,} \\
        \text{Middle Eastern, Latino} \\
    \end{array} 
\right\} \\
\label{eq:gender}
\text{gender} &= \left\{ 
    \begin{array}{l}
        \text{woman, man} \\
    \end{array} 
\right\} 
\end{align}


\subsection{Dataset Demographics}

We use \emph{FaceLib} \cite{FaceLib} to estimate age and gender information from all generated faces. The SynthFace dataset is estimated to be 51.3\% male and 48.7\%; this binary is reductive but useful as a diagnostic. \Cref{fig:SynthFace_age_distribution} details the estimated distribution of ages in the SynthFace dataset, with estimated gender also shown. It is important to document the demographic data of a proposed dataset, as performance can be expected to be worse on those outside of the modelled data distribution. Each generated face reflects data distributions within FLAME, Stable Diffusion, and how these are linked through ControlNet. Through prompting, we aim for a dataset balanced by gender.





  \begin{figure*}
      \centering
      \includegraphics[width=\textwidth]{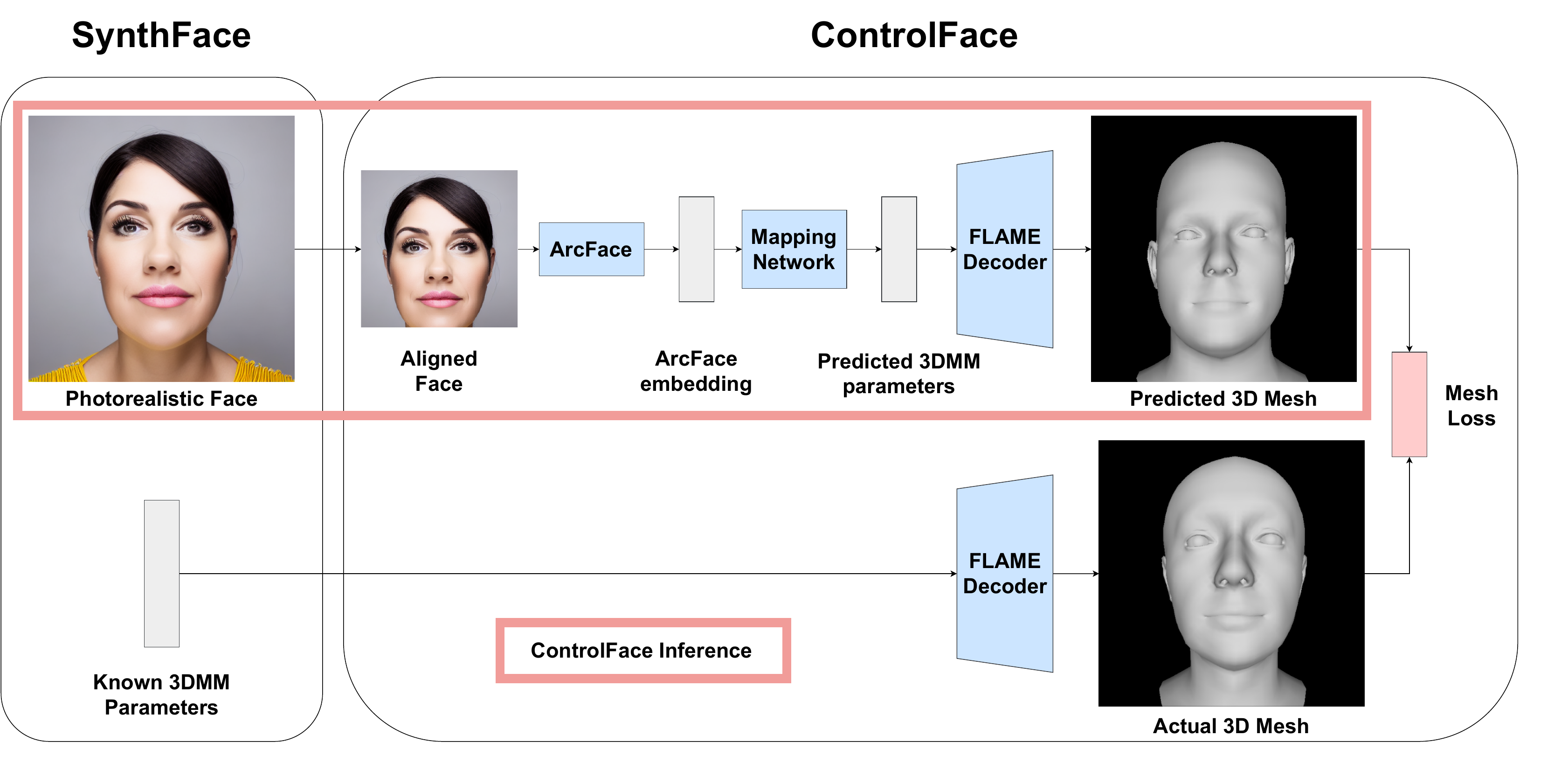}
      \caption{ControlFace training. We train the mapping network within ControlFace on the SynthFace dataset. It is trained to minimise the mesh reconstruction error between a predicted 3D mesh and known 3D mesh for each image in SynthFace. ControlFace at inference is shown outlined. ControlFace accepts an image as input, aligns it, and calculates an ArcFace embedding from this aligned detected face. A mapping network converts this ArcFace embedding to 3DMM parameters. The FLAME decoder generates a full head mesh from these parameters.}
      \label{fig:ControlFaceOverview}
  \end{figure*}

\section{ControlFace for 3D Face Reconstruction}

We introduce \emph{ControlFace}, a deep neural network trained on our new SynthFace dataset. This network aims to disentangle shape from identity and perspective through supervised training on a large dataset that contains multiple identities for the same shape. It accepts an image as input and outputs a shape vector $x \in \mathbb{R}^{300}$ for the FLAME decoder. All architectures, training, and evaluation are implemented using PyTorch \cite{paszke2019pytorch}. \Cref{fig:ControlFaceOverview} shows the training process in full, including the inference pipeline used during model deployment.

\subsection{Training Data}

We use the entirety of the SynthFace dataset as our training data. SynthFace contains 250K images of 10K unique shape identities. A unique shape identity is defined as a unique set of 3DMM parameters. For each of these unique shape identities, we render five depth maps under different perspective projections and five images for each of these depth maps.

\subsection{Pre-processing}

First, faces are detected in each image using \emph{RetinaFace} \cite{deng2020retinaface}. This provides a bounding box used to crop each image and warp it to a frontal pose. The images in SynthFace share a common frontal pose by design. However, this detection and warping step remains crucial. In-the-wild images have various poses which our approach must be able to handle. Next, we use the pretrained ArcFace network as a feature extractor for face description. ArcFace's 512-dimensional output embedding is used as input for a mapping network.

\subsection{Mapping Network}

 We use the same mapping network architecture as presented by Zielonka \etal \cite{zielonka2022towardsMICA}. This network consists of three fully-connected layers followed by a linear output layer. Weights are randomly initialised and we train this network to regress a shape vector $y \in \mathbb{R}^{300}$ from an ArcFace embedding vector $x \in \mathbb{R}^{512}$ . This vector contains coefficients for all 300 identity bases in the FLAME head model.  


\subsection{Training Strategy}

We split the SynthFace dataset into training and validation sets, following an 85/15 split. We train our mapping network on the training set and select the best performing model based on the validation loss; we use early stopping with a patience of 10 to achieve this and run for 100 epochs. 

We use the AdamW optimizer for optimisation with  learning rate $\eta = 1 \times 10^{-5}$ and weight decay $\lambda = 2 \times 10^{-4}$. We use the same optimisation strategy and masked mesh loss function as  Zielonka \etal \cite{zielonka2022towardsMICA}: 
\begin{equation}
L = \sum_{(I, G)} |\kappa_{\text{mask}}(G_{\text{3DMM}}(M(\text{ArcFace}(I))) - G)|,
\end{equation}
which puts emphasis on inner facial regions in reconstruction.
\( \kappa_{\text{mask}} \) is a region-dependent weight mask with values: 150 for the face region, 1 for the back of the head, and 0.1 for the eyes and ears.
This loss is calculated for all pairs of input images, \(I\), and known meshes, \(G\), within SynthFace. \((G_{\text{3DMM}}(M(\text{ArcFace}(I)))\) is the predicted mesh after the image is passed through ArcFace, the mapping network \(M\), and then the FLAME decoder \(G_{\text{3DMM}}\).

\section{Experiments and Evaluation}

\begin{table}[h!]
\centering
\begin{tabular}{lcccc}
\hline
Method &Med.&Mean&Std.&Train\\ 
\hline
Deep3D \cite{deng2019accurate} & 1.286 & 1.864 & 2.361 & \ding{55} \ding{55} \ding{51}\\
DECA (detail) & 1.190 & 1.469 & 1.249 & \ding{55} \ding{55} \ding{51}\\
DECA \cite{feng2021learningdeca} & 1.178 & 1.464 & 1.253 & \ding{55} \ding{55} \ding{51}  \\
AlbedoGAN (detail) & 0.950 & 1.173 & 0.987 & \ding{55} \ding{51} \ding{51} \\
MICA \cite{zielonka2022towardsMICA} & 0.913 & 1.130 & 0.948 & \ding{51} \ding{55} \ding{55} \\
AlbedoGAN \cite{rai2023towards} & 0.903 & 1.122 & 0.957 & \ding{55} \ding{51} \ding{51} \\
ControlFace (ours) & 1.181 & 1.451 & 1.191 & \ding{51} \ding{51} \ding{51}  \\
\hline
\end{tabular}
\caption{Reconstruction error (mm) on the validation set of the NoW benchmark \cite{sanyal2019learningNoW} in non-metrical reconstruction. Comparison results are presented from \cite{rai2023towards}. The final column includes ticks and crosses that indicate  whether the method meets specified criteria. The first element indicates whether supervised training between images and 3DMM parameters is employed. The second and third elements indicate the use of synthetic data: first for 2D images and then for 3D meshes.}
\label{tab:table1}
\end{table}

We test our proposed method against the \emph{NoW} benchmark \cite{sanyal2019learningNoW}. The NoW benchmark consists of 2054 images for 100 identities. It has become the standard benchmark for evaluating 3D face reconstruction from 2D images. These are split into validation and test sets consisting of 20 and 80 identities respectively. For each individual, the dataset includes images under different poses, occlusions, and expressions. We use the publicly available validation set of NoW for evaluation. First, a rigid alignment of the predicted meshes to the scans is performed using key facial landmarks. Then the scan-to-mesh distance between the predicted mesh and scan is performed for each vertex. The mean, median, and standard deviations of these distances is computed across all images within the validation set. \Cref{tab:table1} shows a comparison of our ControlFace approach with current state-of-the-art methods. All methods presented use supervised or self-supervised learning.

Our results are competitive with the current state-of-the-art in 3D face shape estimation - crucially, without requiring any ground truth 3D shape data. We achieve this by introducing a novel method for large dataset generation for 3D face shape estimation. Our work with ControlFace demonstrates that supervised training on this dataset leads to accurate 3D face shape estimation.

Our work is easily extensible. A longer generation time can lead to a larger dataset and improvements in 2D and 3D generative model capabilities can directly feed into future work. We believe this will enable future versions of SynthFace to close the performance gap with methods such as MICA and AlbedoGAN. Datasets for specific use cases, be that large pose variations or expressions, can be created by updating parameters in our generation code. 

In unifying existing 3D face datasets, MICA reaches a natural limit in supervised learning on existing data sources. This is where the opportunity for synthesised approaches such as SynthFace lies. SynthFace can scale beyond this natural limit in real paired data.

\section{Limitations and Future Work}

The current iteration of SynthFace exclusively models variations in shape, leaving out expressive variations. Consequently, ControlFace solely focuses on shape prediction. It may be beneficial for future research to include varying expressions within the dataset or to devise a separate network to model these variations independently.

Our method employs ArcFace to generate a facial identity descriptor, which serves as the input to our mapping network. Importantly, this is an identity embedding and not a shape embedding. We make the assumption that the ArcFace-learned identity encompasses shape and that our mapping network can extract shape from this. Future research should explore retraining ArcFace or similar networks to more specifically extract shape information. Furthermore, the embedding network could be removed entirely, replacing it with a single network that learns to map images to 3DMM parameters in a supervised manner.

We utilise individual depth maps derived from a 3D face model to condition Stable Diffusion. Our knowledge of the full 3D geometry could be utilised further to improve the conditioned image. This could involve multi-image or even multi-modal conditioning to allow for even greater shape consistency between the 3D model and the generated 2D image.


We must also consider the ethical implications of our work. SynthFace is designed to be balanced by race and gender. However, despite best efforts, we understand that these selected subgroups do not cover every identity or individual. We also recognise that we use a deep-learning based age and gender estimator for this analysis which itself may well be biased. 


We agree with Buolamwini and Gebru \cite{buolamwini2018gender} in their proposal for intersectional error analysis using a benchmark balanced by gender and skin colour. In creating a dataset with equal representation of subgroups by race and gender, we are the first to enable this form of intersectional analysis for 3D face reconstruction. Further work should consider evaluating state-of-the-art reconstruction methods using this approach.


Generative models like Stable Diffusion require extensive datasets for training that typically rely on publicly available data. Consequently, there's a likelihood that individuals' data has been used without their explicit consent. This raises clear ethical and legal concerns, particularly for models deployed in the real world.

Accurate 3D face shape estimation finds  application in areas such as prosthesis design, yet it can also be utilised for malevolent purposes, including deepfake creation and mass surveillance. These potential misuses must be considered during model development and deployment and weighed against potential benefits.

\section{Conclusion}

We have addressed a key challenge in 3D face shape estimation by proposing a method for generating a large-scale dataset for supervised training. Our method combines existing 2D and 3D generative models to produce photorealistic images, balanced by race and gender, with corresponding 3DMM parameters. The resulting dataset, SynthFace, is the largest dataset of its kind and offers unique opportunities to disentangle shape from identity for accurate 3D face reconstruction. The complete SynthFace dataset will be made publicly available upon publication.

Our 3D face reconstruction results prove competitive with the existing state-of-the-art. Notably, our technique does not rely on ground truth 3D shape data. Unlike previous methods, ours is easily extensible, computationally inexpensive, and produces photorealistic face images. It further addresses race and gender bias in existing computer vision datasets, providing a dataset balanced by race and gender that can be used for training as well as model audit. We see this approach to solving 3D problems by using conditioned 2D diffusion models to hold great potential, particularly as existing 3D face datasets reach their limit for supervised learning. 



We expect improvements in image generation, 3D face models, and conditioning networks to all improve the accuracy of this method for 3D face reconstruction; our work provides a clear path for continuous improvement. We believe this work will form the basis of a number of exciting future developments in this domain.

{\small
\bibliographystyle{ieee_fullname}
\bibliography{egbib}
}

\end{document}